\documentclass[12 pt,letterpaper]{article}
\usepackage{named}
\usepackage{graphicx} %
\usepackage{amsmath} %
\usepackage{amsthm} %
\usepackage[hyphens]{url}  %
\makeatletter
\def\url@leostyle{%
  \@ifundefined{selectfont}{\def\UrlFont{\sf}}{\def\UrlFont{\small\ttfamily}}}
\makeatother
\title{An Effective Procedure for Computing ``Uncomputable" Functions\thanks{This work is licensed under the Creative Commons Attribution-NonCommercial-NoDerivs 3.0 Unported License (see http://creativecommons.org/licenses/by-nc-nd/3.0/).}
}
\author{Kurt Ammon\footnote{Correspondence to p.k.ammon [at] cproc [period] org. Comments are welcome.}} 
\date{} %

\newtheorem{rl}{Rule}
\newcommand{\br}{\begin{rl}\rm}
\newcommand{\er}{\end{rl}}

\newcommand{\bt}{\vspace{-.1cm}\begin{tabular}}
\newcommand{\et}{\end{tabular}\vspace{-.1cm}}

\newcommand{\bc}{\begin{center}}
\newcommand{\ec}{\end{center}}
\newcommand{\bqu}{\begin{quote}}
\newcommand{\equ}{\end{quote}}
\newcommand{\bci}{\begin{center} \begin{minipage}{7.5cm} }
\newcommand{\eci}{\end{minipage} \end{center} }
\newcommand{\be}{\begin{enumerate}}
\newcommand{\ee}{\end{enumerate}}
\newcommand{\beq}{\begin{equation}}
\newcommand{\eeq}{\end{equation}}
\newcommand{\barr}{\begin{array}}
\newcommand{\earr}{\end{array}}

\newtheorem{dfn}{Definition}
\newcommand{\bdf}{\begin{dfn}\rm}  
\newcommand{\edf}{\end{dfn}}  
\newtheorem{thm}{Theorem}
\newcommand{\bth}{\begin{thm}\rm}  
\newcommand{\eth}{\end{thm}}  
\newtheorem{exm}{Example}
\newcommand{\bex}{\begin{exm}\rm}  
\newcommand{\eex}{\end{exm}}  
\newcommand{\bpr}{\begin{proof}}  
\newcommand{\epr}{\end{proof}}  
\newtheorem*{hyp}{Hypothesis}
\newcommand{\bhp}{\begin{hyp}\rm}  
\newcommand{\ehp}{\end{hyp}}
\newtheorem*{pri}{Existence Principle}
\newcommand{\bpri}{\begin{pri}\rm}
\newcommand{\epri}{\end{pri}}
\newcommand{\mc}{\multicolumn}  
\begin{document} %
\maketitle
\begin{abstract}
We give an effective procedure that 
produces a natural number in its output 
from any natural number in its input, 
that is, it computes a total function. 
The elementary operations of the procedure are Turing-computable.
The procedure has a second input which can contain
the G\"odel number of any 
Turing-computable total function whose range is a subset 
of the set
of the G\"odel numbers of all Turing-computable total 
functions.
We prove that the second input cannot be
set to the G\"odel number of any Turing-computable function that computes the output from any natural number in 
its first input. In this sense, there is no Turing program that computes 
the output from its first input. 
The procedure is used to define
creative procedures which compute functions that are not Turing-computable.
We argue that creative procedures model an aspect of reasoning that 
cannot be modeled by Turing machines.
\end{abstract}

\section{Introduction}  \label{SIntr}
There are doubts whether Turing machines can capture 
all reasoning processes.
For example, Turing \shortcite[pp.\ 200, 215]{Turing39} writes: %
\bqu
G\"odel's theorem shows that such a system [intellectually satisfying system of logical inference] cannot be wholly mechanical ...
The necessity for using the intuition is then [by the introduction of a formal logic] 
greatly reduced by setting down formal rules 
for carrying out inferences which are intuitively valid. ...
In pre-G\"odel times it was thought by some that it would probably be possible
to carry this programme to such a point 
that all the intuitive judgments
of mathematics could be replaced by a finite number of these rules.
The necessity for intuition would then be entirely eliminated.
\equ
Thus, Turing \shortcite{Turing39} 
interprets G\"odel's theorem in the sense 
that an ``intellectually satisfying system of logical inference 
... cannot be wholly mechanical" 
and
``all the intuitive judgments
of mathematics" cannot ``be replaced by a finite number of these rules", that is, 
``formal rules for carrying out inferences which are intuitively valid".

Turing  \shortcite[p.\ 231]{Turing36} restricts his machines to a finite number of {\em m}-configurations (machine configurations)
which are called ``states of mind" 
in his \shortcite[pp.\ 249-250]{Turing36} substantiation of the thesis
 that his machines 
can compute ``all numbers which would naturally be regarded as computable" (Turing's thesis).
Turing \shortcite[pp.\ \ 249-250]{Turing36} 
supposes that the ``number of states of mind" is finite
because some of them ``will be confused" if ``we admitted an infinity of states of mind"
(see Kleene \shortcite[pp.\ 376-377]{Kleene52}). 
G\"odel \shortcite[p.\ 306]{Goedel72a}
regards the restriction to a finite number of states as a ``philosophical error
in Turing's work"
and points out
that ``mental procedures" may ``go beyond mechanical procedures".
G\"odel \shortcite[p.\ 306]{Goedel72a} writes:
\bqu
What Turing disregards completely is the fact that
{\em mind, in its use, is not static, but constantly developing}, ... 
There may exist systematic methods of actualizing this development, which could form part of the procedure.
Therefore, 
although at each stage the number and precision of the abstract terms at our disposal may be {\em finite}, both (and, therefore, also Turing's number of {\em distinguishable states of mind}) may {\em converge toward infinity}
 in the course of the application of the procedure.
\equ
Thus, G\"odel discusses the possible existence of ``mental" procedures that cannot be modeled by any Turing machine
which is restricted to a finite number of %
"states of mind" in 
Turing's \shortcite[pp.\ \ 249-250]{Turing36} substantiation of his thesis. 

Referring to his own form of G\"odel's incompleteness theorem,
Post \shortcite[p.\ 295]{Post44} %
writes:
\bqu
The conclusion is unescapable that even for such a
fixed, well defined body of mathematical propositions, 
{\em mathematical thinking is, and must remain, essentially creative}.
\equ
G\"odel's \shortcite[p.\ 5]{Goedel31} incompleteness theorem is based on ``Principia Mathematica and related systems".
Referring to his result that 
there is no finite method deciding %
whether a sequence is generated by the operations of a normal system
 Post \shortcite[pp.\ 407-408]{Post65}
writes:
\bqu
... the analysis ... is fundamentally weak in its reliance on the logic
of Principia Mathematica 
...
But for full generality a complete analysis would have to be made
of all possible ways in which the human mind could set up
finite processes for generating sequences.
\equ
 Davis \shortcite[p.\ 21]{Davis82} writes that Post ``evidently felt that 
the very incompleteness of 'Principia Mathematica' ...
undermined its suitability as a basis for such an analysis."

Turing's, G\"odel's and Post's remarks suggest the possible existence of "mental" procedures that cannot be reduced to 
Turing machines. 
Section \ref{SProc} introduces an effective procedure that 
computes a total function. 
We prove that a second input of the procedure cannot be
set to the G\"odel number of any Turing program that computes the output from any natural number in 
its first input. %
The procedure concerns the question whether every procedure used in an "intelligent" system can 
be modeled by a Turing machine. In particular, it concerns the question whether 
an "intelligent" system can model all its own functions by a Turing machine.

Section \ref{SDiscussion} uses the procedure in Section \ref{SProc} to define
creative procedures which compute functions that are not Turing-computable.
We argue that creative procedures capture an aspect of 
G\"odel's ``mental procedures" which are not Turing-computable.
In Section \ref{SRelated} we discuss Church's thesis.

\section{Procedure}\label{SProc}
For his own form of G\"odel's incompleteness theorem Post \shortcite{Post44}
introduced %
creative sets %
whose definition implicitly refers to productive 
functions.\footnote{The term {\it productive} is due to Dekker \shortcite{Dekker55}.}
We regard the existence of productive functions as a key to the phenomenon of incompleteness. %

We deal with natural numbers, sets of natural numbers, and functions from natural numbers to natural numbers.
We use the following terminology and notations:
We write ${\it domain} \, \varphi$ for the {\it domain} of a function $\varphi$ 
and ${\it range} \, \varphi$ for the {\it range} of $\varphi$.
A function is called {\it partial} if its domain is a subset of the set all natural numbers.
A function is called {\it total} if its domain is the set of all natural numbers.
If $\varphi$ is a partial function of natural numbers, we say that $\varphi$ is {\it defined}
at the natural number $x$ if $x \in {\it domain} \, \varphi$.

Because Turing machines are represented as finite sets of instructions, that is, as finite sequences of a
fixed finite number of symbols, 
it is possible to list the sets of instructions of all Turing machines by an algorithm, for example, 
in ascending length according to the number of symbols that a set of instructions contains.
We follow Rogers \shortcite[p.\ 21]{Rogers87} and keep such a listing fixed for the remainder of this article:
\bdf\label{dPi}
The {\it Turing program}
$P_i$ is the set of instructions of a Turing machine associated with the natural number $i$ in a fixed listing of 
the sets of instructions of all Turing machines.
$i$ is called the {\it index} or {\it G\"odel number} of $P_i$.
$\varphi_i$ is the partial function determined by $P_i$.
$i$ is also called the {\it index} or {\it G\"odel number} of $\varphi_i$.
\edf
The listing gives an algorithm for generating $P_i$ from any natural number $i$ and another algorithm for generating 
a natural number $i$ from the set of instructions $P$ of any Turing machine such that $P$ is $P_i$.
The two algorithms can be encoded as ordinary computer programs.

A set of natural numbers is recursively enumerable if there is an algorithm for enumerating its members.
 A precise definition is \cite[p.\ 58]{Rogers87}:

\bdf\label{dre}
A set of natural numbers is {\em recursively enumerable} if it is empty or the range of a Turing-computable total function. 
\edf

A set is called productive if there is a mechanical procedure (algorithm) which, given any recursively enumerable subset, produces a member of the set that is not contained in the given subset. A precise definition
is: 
\bdf\label{dprod}
A set $S$ of natural numbers is {\em productive} if there is a Turing-computable partial function $\psi$ such that, given any 
 total function $\varphi_j$ whose range is a subset of $S$,
the value
 $\psi(j)$  is defined and contained in $S$ but not in the range of $\varphi_j$:

\bc 
\hspace*{-1.2cm}$(\forall j)[\,\varphi_j \, \text{\it total} \;\, \& \;\, \text{\it range}\, \varphi_j 
	\subseteq S \;\Rightarrow\;$ \\ [.01cm]
\hspace*{2cm}$[\, \psi(j)\, \text{\it defined} \;\, \& \;\, \psi(j) \in S\!-\! 
	\text{\it range}\, \varphi_j \, ]] $
\ec
The partial function $\psi$ is called a {\em productive partial function for} $S$.
\edf
Definition \ref{dprod} is equivalent to the definition of productive sets in Rogers
\shortcite[p.\ 84, see p.\ 90]{Rogers87} because of basic theorems such as Rogers
\shortcite[p.\ 60, Theorem V, and p.\ 61, Corollary V(b)]{Rogers87}.

The following theorem %
states that 
the set
$\{ i  |\varphi_i \, {\it total} \}$ of the G\"odel numbers  $i$ of all 
(Turing-computable) total 
functions $ \varphi_i$ is productive. 
According to Definition \ref{dprod}, this means that
there is a Turing-computable partial function $\psi$, which is called
productive,
such that,
given any 
(Turing-computable) total function $\varphi_j$ 
with $range \, \varphi_j \subseteq \{ i  |\varphi_i \, {\it total} \}$, 
the value
 $\psi(j)$  is defined and contained in $\{ i  |\varphi_i \, {\it total} \}$ but not in the range of $\varphi_j$.

\begin{thm}\label{tprod}
The set
$\{ i  |\varphi_i \, {\it total} \}$ of the G\"odel numbers  $i$ of all total 
functions $ \varphi_i$ is productive. %
\end{thm}
\bpr
Let $j$ be any natural number such that $\varphi_j$ is a total function with 
$range \, \varphi_j \subseteq \{ i  |\varphi_i \, {\it total} \}$, 
that is, the range of $\varphi_j$ is a set of G\"odel numbers %
of Turing-computable total functions.
Thus, $\varphi_{\varphi_j(n)}$
is a Turing-computable total function for any natural number $n$.
Using Cantor's diagonal method we define a new function
$\delta_j$ %
 by 
\beq
 \delta_j(n)=\varphi_{\varphi_j(n)}(n)+1 \label{delta}
\eeq 
for all natural numbers $n$.  
Obviously, $\delta_j$ is a %
total function because
 $\varphi_{\varphi_j(n)}$
is a %
total function for any natural number $n$.

In order to prove the theorem,
we construct a Turing-computable procedure computing
a %
partial 
function $\psi$ whose input is
 any natural number $j$ 
satisfying the properties 
that 
$\varphi_j$ is a total %
and
$range \, \varphi_j \subseteq \{ i  |\varphi_i \, {\it total} \}$. 
Such a natural number $j$ is 
given at the beginning of the proof and is
used in the definition 
 of the function $\delta_j$ in (\ref{delta}).
The output $\psi(j)$  of the %
 function $\psi$
for the input $j$ %
is the G\"odel number $\psi(j)$ of a Turing program $P_{\psi(j)}$
computing the function $\delta_j$ in (\ref{delta}).
The  G\"odel number $\psi(j)$ is constructed as follows:
According to Definition \ref{dPi}, the function $\varphi_j$ in (\ref{delta}) is computed by the Turing 
program $P_j$ and %
the function  $\varphi_{\varphi_j(n)}$ in (\ref{delta}) is computed by the Turing program 
$P_{\varphi_j(n)}$. %
The expression $\varphi_{\varphi_j(n)}(n)+1$ in (\ref{delta}) can be regarded as a %
pseudo-code for a Turing program $P_{\psi(j)}$ which computes the function $\delta_j$ in (\ref{delta}) 
and can be constructed from the Turing programs
 $P_j$ and $P_{\varphi_j(n)}$, where $\varphi_j(n)$ is the result of applying $P_j$
to $n$.
The construction of  $P_{\psi(j)}$ from  $P_j$ and $P_{\varphi_j(n)}$
can be achieved by a Turing-computable procedure that is independent of
the G\"odel number $j$
because the states of $P_j$ and $P_{\varphi_j(n)}$ 
(called $m$-configurations, that is, machine configurations, in Turing \shortcite{Turing36}
and internal states in Rogers \shortcite[p.\ 13]{Rogers87})
can be renamed such that 
$P_j$ and $P_{\varphi_j(n)}$ only contain different states 
and can thus be used as components of $P_{\psi(j)}$.
Because of Definition \ref{dPi}, the 
G\"odel number $\psi(j)$ of $P_{\psi(j)}$ can be generated from $P_{\psi(j)}$
by a Turing-computable procedure.
From the Turing-computable procedures producing
${\psi(j)}$ from $P_{\psi(j)}$ and
$P_{\psi(j)}$ from  $P_j$ and $P_{\varphi_j(n)}$ 
we can construct a Turing-computable procedure
that computes $\psi(j)$ from the G\"odel  number $j$ of any total function $\varphi_j$
with $range \, \varphi_j \subseteq \{ i  |\varphi_i \, {\it total} \}$.
Therefore, $\psi$ is a Turing-computable partial function
that computes the G\"odel number $\psi(j)$ of a 
Turing program $P_{\psi(j)}$ computing the
total function $\delta_j$
from the G\"odel  number $j$ of any total function $\varphi_j$
with $range \, \varphi_j \subseteq \{ i  |\varphi_i \, {\it total} \}$
which was given at the beginning of the proof.

Because of (\ref{delta})
the function $\delta_j$
 is different from 
$\varphi_{\varphi_j(n)}$ for all natural numbers $n$.
Because the Turing program $P_{\psi(j)}$,
which computes $\varphi_{\psi(j)}$ according to Definition \ref{dPi},
also computes the function $\delta_j$ in (\ref{delta}),
$\delta_j(n)=\varphi_{\psi(j)}(n)$ for all natural numbers $n$.
Therefore, the total function $\varphi_{\psi(j)}$ 
 is different from 
$\varphi_{\varphi_j(n)}$ for all natural numbers $n$.
In particular, 
 $\psi(j)$ 
 is different from 
$\varphi_j(n)$
for all natural numbers $n$
because different functions
$\varphi_{\psi(j)}$ and
$\varphi_{\varphi_j(n)}$ 
cannot have the same G\"odel number according to Definition \ref{dPi}.
This implies that
$\psi(j)$ is not contained in $range \, \varphi_j$,
that is, the range of $\varphi_j$.

Therefore, $\psi$ is a productive partial function
for the set
$\{ i  |\varphi_i \, {\it total} \}$ of the G\"odel numbers  $i$ of all total 
functions $ \varphi_i$. In view of  Definition \ref{dprod} this completes the proof of the
 theorem.

\epr

Rogers \shortcite[p.\ 84, {\it Example} 2]{Rogers87} suggests another proof 
that the set $\{ i  |\varphi_i \, {\it total} \}$ is productive.

\bdf
A recursively enumerable set $S$ of natural numbers is {\em creative} if its complement is productive.
\edf

Referring to G\"odel's \shortcite{Goedel31} incompleteness theorem,
Rogers \shortcite[pp.\  97-98]{Rogers87} states that
[the G\"odel numbers of]
 the provable well-formed formulas of Peano arithmetic form a creative set,
[the G\"odel numbers of]
the unprovable well-formed formulas form a productive set,
and 
[the G\"odel numbers of]
the true
well-formed formulas of elementary arithmetic form a productive set.
Rogers \shortcite[p.\  98]{Rogers87} writes:
\bqu
... no axiomatization of mathematics can exactly capture
all true statements in elementary arithmetic;
and from any axiomatization which yields only
true statements in elementary arithmetic, a  new true
statement can be found not provable in that axiomatization.

... Post believed that such facts manifest an essentially creative 
quality of mathematics; hence the name {\it creative set}.
\equ
Soare \shortcite[p.\ 1151]{Soare78} writes:
\bqu
Such r.e.\ [recursively enumerable] sets were called creative by Post ... because their existence 
... %
implies the impossibility of mechanically listing all
statements true in such a fragment [fragment of mathematics as
elementary number theory].
\equ

The construction of undecidable formulas of formal systems in G\"odel's \shortcite{Goedel31} incompleteness theorem can be represented by an algorithm. 
Productive functions (see Definition \ref{dprod}), which are Turing-computable, 
correspond to a formal abstraction of this algorithm. 
Thus, the input of productive functions corresponds to formal systems.
This suggests that formal systems, in particular, 
Turing programs, cannot refer to themselves, that is, they cannot capture their own {\em existence}. 
Otherwise, %
the application of productive functions, 
which correspond to the construction of the undecidable propositions,
could be used to ``overcome"
incompleteness as described below.

We use any productive function $\psi$ %
for the set of the G\"odel numbers  
of all Turing-computable total functions
in Theorem \ref{tprod} to define a procedure $Q$ which computes
the output $\omega(x)$  of a total function $\omega$ for any natural number $x$ in its input
and prove that function $\omega$ is not Turing-computable.
The procedure 
$Q$ has a {\em second input} $j$ which is in the domain of $\psi$
and contains the G\"odel number 
of a  Turing program representing an {\em existing\/} formal system.
The function $\omega$ is not Turing-computable
because
G\"odel numbers $\psi(j)$,  where
$j$ is an  {\em existing\/} G\"odel number in the domain of $\psi$, are contained in the output of
the procedure $Q$, that is, the output of $\omega$.
Roughly speaking, there is no G\"odel number $j$ of a Turing program $P_j$ %
 generating all G\"odel numbers of Turing-computable total functions
that a human (or a ``machine") generates
if the human (or the ``machine") 
applies %
$\psi$ 
 to the {\em existing\/} G\"odel number $j$ and thus produces the G\"odel number $\psi(j)$  of a Turing-computable total function
that is not generated by $P_j$.

\begin{table}
\bc
\begin{math}
\barr {rlll}
1. &    \mc {3}{l}{l := \text{\it least } \{ n \in {\bf N} \, | \ n \notin \text{\it domain } \alpha \} ; } \\
2.& \text{\it if} &  \mc {2}{l}{\text{\it is-set } (j)} \\
3. & &  \mc {2}{l}{\hspace*{.1cm}\text{\it then } \;\; \alpha := \alpha \cup \{ (l,  \psi(j)) \} ; }\\ 
4.& \text{\it if} &  \mc {2}{l}{\text{\it is-not-set } (x)} \\
5. & &  \mc {2}{l}{\hspace*{.1cm}\text{\it then } \;\; \text{\it return } 1\, ; }\\ 
6.& \text{\it if} &  \mc {2}{l}{x \in \text{\it domain } \alpha} \\
7. & &  \mc {2}{l}{\hspace*{.1cm}\text{\it then } \;\; \text{\it return } \alpha(x); }\\ %
8. &   \mc {3}{l}{\alpha := \alpha \cup \{ (x,  c) \}; }\\
9. &   \mc {3}{l}{\text{\it return } \alpha(x);}\\
\earr
\end{math}
\ec
\vspace*{-.2cm}
\caption{\it  Procedure $Q$ with two inputs $x$ and $j$ and a global variable $\alpha$
for computing $\omega(x)$\label{computing}}
\end{table}   

\bdf \label{dproc}
The procedure $Q$, whose  pseudo-code is given in Table \ref{computing}, has two input variables $x$ and $j$.
The variable $x$ is not set or set to any natural number, that is,
$x$ has no value or the value of $x$ is any natural number. 
The variable $j$ is not set or set to the G\"odel number  of any 
Turing-computable total function whose range is a subset 
of the set
$\{ i  |\varphi_i \, {\it total} \}$ of the G\"odel numbers  $i$ of all total 
functions $ \varphi_i$.

The global variable $\alpha$ in the procedure $Q$ in Table \ref{computing}
is set to the empty set $\emptyset$ 
before the first execution of the procedure $Q$. 
The variable $\alpha$
is a function which is represented as a set of input-output pairs
$(x,y)$, that is, $(x,y) \in \alpha$ means $\alpha(x) = y$ in ordinary notation.
The variable $\alpha$ is only changed by the 
procedure $Q$ itself.

The first line of the procedure $Q$ in Table \ref{computing}
sets the variable $l$ to the least natural number $n\in {\bf N}$
that is not contained in the domain of the function $\alpha$
which was set to the empty set $\emptyset$ before the first execution of the procedure $Q$.
The function $\psi$ in the third line of the procedure $Q$ 
is a productive partial function for 
the productive set
$\{ i  |\varphi_i \, {\it total} \}$ of the G\"odel numbers  $i$ of all total 
functions $ \varphi_i$. Such a productive partial function $\psi$ exists according to
Theorem \ref{tprod} (see Definition \ref{dprod}).
If the second input variable $j$ is set (has a value), 
that is,
the condition in the second line of the procedure $Q$ is satisfied,
the third line adds the input-output pair $(l,\psi(j))$ to the function $\alpha$,
which is represented as a set of input-output pairs. 
If the first input variable $x$ is not set (has no value), 
that is,
the condition in the fourth line of $Q$ is satisfied,
the fifth line returns $1$ as the output of the procedure $Q$.
If the condition $x \in \text{\it domain } \alpha$ in the sixth line is satisfied,
the procedure $Q$ returns $\alpha(x)$ as its output in the seventh line.
Otherwise, the eighth line
adds the input-output pair 
$(x, c)$ %
to the function $\alpha$,
where the constant $c$ %
is the G\"odel number of any fixed Turing-computable total function.
Finally, the ninth line 
returns $\alpha(x)$,  which is equal to $c$ because of the eighth line, 
as the output of the procedure $Q$.
\edf

\bth \label{tproc1}
All elementary operations of the procedure $Q$ 
in Definition \ref{dproc} and Table \ref{computing} 
are Turing-computable.
\eth
\bpr
The least natural number $n\in {\bf N}$
that is not contained in the domain of the function $\alpha$
in the first line in Table \ref{computing}
is Turing-computable because $\alpha$ is a finite set of input-output pairs at every point in time.
The expression $\psi(j)$ in the second line in Table \ref{computing}
is Turing-computable
because $j$ is
the G\"odel number  of any 
Turing-computable total function whose range is a subset 
of the set
$\{ i  |\varphi_i \, {\it total} \}$ of the G\"odel numbers  $i$ of all total 
functions $ \varphi_i$ according to Definition \ref{dproc},
$j$ is in the domain of $\psi$ according to Definition \ref{dprod} and Theorem \ref{tprod}, 
and $\psi$ is Turing-computable according to Definition \ref{dprod} and Theorem \ref{tprod}.
Obviously, the other elementary operations in the procedure $Q$ %
are also Turing-computable.
\epr

\bth \label{tproc2}
The procedure $Q$ in Definition \ref{dproc} and Table \ref{computing} 
computes a total function whose input is any natural number $x$ in the first input
of %
$Q$ 
and
whose output is the output of $Q$, where the second input variable $j$ of $Q$ 
is not set or set and may be changed at any time according to Definition \ref{dproc}.
\eth
\bpr
The value $\alpha(x)$, where $x$ is any natural number, in the seventh and the ninth line in Table \ref{computing} is uniquely determined by 
the set $\alpha$ of input-output pairs representing the function $\alpha$:
Let  $x$, $y_1$, and $y_2$ %
be any natural numbers with
$(x,y_1) \in \alpha$ and $(x,y_2) \in \alpha$. This implies $y_1=y_2$ because of the
construction of $\alpha$ in the first and the third line %
and the extension of $\alpha$ in the eighth line is only used if 
$x \notin \text{\it domain } \alpha$. 
Thus, the value $\alpha(x)$ is uniquely determined.
Therefore, the value $\omega(x)$ is uniquely determined because 
$\omega(x)=\alpha(x)$ for any natural number
\mbox{$x \in \text{\it domain } \alpha$}.

The procedure $Q$ in Table~\ref{computing} computes the value $\omega(x)$  of the function $\omega$ for any natural number $x$ because the input-output pair $(x,c)$ is added to $\alpha$ in the eighth line 
if \mbox{$x \notin \text{\it domain } \alpha$}.
Therefore, the domain of the function $\omega$, which is computed by the procedure $Q$,
is the set of all natural numbers, that is, $\omega$ is a total function.
\epr

\bdf \label{domega}
We write $\omega$ for the total function
computed by the procedure $Q$ according to Theorems \ref{tproc1} and \ref{tproc2}. 
\edf

\bth \label{trange}
The range of the total function $\omega$ in Definition \ref{domega} is a subset of the
set $\{ i  |\varphi_i \, {\it total} \}$ of all Turing-computable total functions.
\eth
\bpr
According to Definition \ref{dproc}
the second input $j$ of the procedure $Q$ in Table \ref{computing}
is not set or set to the G\"odel number  of any 
Turing-computable total function whose range is a subset 
of %
$\{ i  |\varphi_i \, {\it total} \}$. %
Thus, the value $\psi(j)$ in the third step of $Q$
is the G\"odel number of a 
Turing-computable total function.
In the third step of $Q$, the value $\psi(j)$ is used as an output of
the function $\alpha$.
According to Definition \ref{dproc}, the value $c$ 
in the eighth step of $Q$
is the G\"odel number of a fixed Turing-computable total function.
In the eighth step of $Q$, the value $c$ is used as an output of
the function $\alpha$.
Thus, the range of $\alpha$
is a subset of $\{ i  |\varphi_i \, {\it total} \}$.
This implies that
 range of $\omega$
is a subset of $\{ i  |\varphi_i \, {\it total} \}$
because any member in the range of $\omega$
is contained in the range of $\alpha$
according to the seventh and the ninth step
of $Q$.
\epr

The following examples illustrate the computation
of the function $\omega$ by the procedure $Q$.

\bex
According to Definition \ref{dproc}
the global variable $\alpha$ in the procedure $Q$ in Table \ref{computing} is set to the empty set
 $\emptyset$ before the first execution of the procedure $Q$. 

In order to compute, for example, the output $\omega(1)$ of the input $1$ we apply the procedure $Q$ to the value $1$ of its first 
input variable $x$. According to the eighth line in Table \ref{computing}, the input-output pair
$(1,c)$ is added to the function $\alpha$ which is represented as a set of input-output pairs. 
According to the ninth line in Table \ref{computing}, the procedure $Q$ returns
\beq
\omega(1)=\alpha(1)=c
\eeq
as the output  $\omega(1)$ of the input $1$.

In order to compute the output  $\omega(5)$ of the input $5$ we apply the procedure $Q$ to the value $5$ of its first 
input variable $x$. Thus, the input-output pair
$(5,c)$ is added to the function $\alpha$ and the procedure $Q$ returns
\beq
\omega(5)=\alpha(5)=c
\eeq
as the output $\omega(5)$ of the input $5$.

Let $j_1$ be the G\"odel number of 
any Turing-com\-putable total function $\varphi_{j_1}$
whose range is a subset of $\{ i  |\varphi_i \, {\it total} \}$.
We apply the procedure $Q$ 
to the value $j_1$ of its second 
input variable $j$.
Because the value of $j$ is set,
the  third line in Table \ref{computing}
adds the input-output pair $(2,\psi(j_1))$
 to the function $\alpha$,
where 
$2= \text{\it least } \{ n \in {\bf N} \, | \ n \notin \text{\it domain } \alpha \} $
according to the first line %
in Table~\ref{computing}.

In order to compute the output  $\omega(2)$ of the input 
$2$ we apply the procedure $Q$ to the value $2$ of its first 
input variable $x$. 
Because 
$2 \in \text{\it domain } \alpha$, 
the procedure $Q$ returns
\beq
\omega(2)=\alpha(2)=\psi(j_1)
\eeq
as the output $\omega(2)$ of the input $2$ according
to the seventh line in Table \ref{computing}.
\eex

\bex \label{Qexseq}
We set
the global variable $\alpha$ in the procedure $Q$ in Table \ref{computing} to the empty set
 $\emptyset$ before the first execution of $Q$. 

Let $j_1$ be the G\"odel number  of %
any Turing-com\-putable total function $\varphi_{j_1}$
whose range is a subset of $\{ i  |\varphi_i \, {\it total} \}$.
We apply the procedure $Q$ 
to the value $j_1$ of its second 
input variable $j$. Thus,
the  third line in Table \ref{computing}
adds the input-output pair $(1,\psi(j_1))$
 to the function $\alpha$,
where 1 is the least natural number that is not contained in the domain of $\alpha$
according to the first line %
in Table~\ref{computing}.
Then, the computation of the output  $\omega(1)$ yields
\beq
\omega(1)=\alpha(1)=\psi(j_1).
\eeq

We construct a sequence of
 G\"odel numbers   $j_2$, $j_3$, ... of %
Turing-computable total functions 
$\varphi_{j_2}$, $\varphi_{j_3}$, ... as follows:
Let $k > 1$ be any natural number.
We define a function $\varphi_{j_k}$
by 
$\varphi_{j_k}(1)=\psi(j_{k-1})$ %
 and
\beq
\varphi_{j_k}(x)=\varphi_{j_{k-1}}(x-1) \label{varphijkx}
\eeq
for all natural numbers $x > 1$.
Thus, the range of $\varphi_{j_k}$ contains $\psi(j_{k-1})$ and $\varphi_{j_{k-1}}(x)$ for any natural number $x$.
Obviously, $\varphi_{j_k}$ is a Turing-computable total
function
whose range is a subset of $\{ i  |\varphi_i \, {\it total} \}$.

The application the procedure $Q$ 
to the value $j_k$ of its second 
input variable $j$
adds the input-output pair $(k,\psi(j_k))$ to the function $\alpha$.
Therefore, the computation of the output  $\omega(k)$ yields
\beq
\omega(k)=\alpha(k)=\psi(j_k)
\eeq
for any natural number $k$.

\eex

The procedure $Q$ in Table~\ref{computing} contains two input variables
$x$ and $j$.
If $Q$ is applied to any natural number $x$, $Q$ produces an output $\omega(x)$,
that is, $Q$ computes a total function.
If $\omega$ were Turing-computable, there would be a Turing program computing $\omega$.
The proof of the following theorem shows
that the input variable 
$j$ of $Q$ cannot be set to the G\"odel number of this Turing program, that is, $P_j$,
because the productive function $\psi$ in $Q$ is applied to $j$
and $\psi(j)$ is used in the output of $\omega$ such that
Theorem \ref{tprod} precludes that $P_j$ computes $\omega$.
The condition in the following theorem 
entails that $\psi(j)$ %
is in the output of the function $\omega$ according to 
the third line of the procedure $Q$. Therefore, 
the Turing program $P_j$ does not compute $\omega$ 
because $\psi(j)$ is not contained in the output of $P_j$ %
according to Theorem \ref{tprod}.
An %
explanation is that the G\"odel number $\psi(j)$ can be used to construct
a more powerful Turing program that produces $\psi(j)$ and all 
G\"odel numbers in the output of $P_j$.
Thus, $Q$ can be used to construct more and more powerful Turing programs.

\bth \label{tomega}
There is no G\"odel number $j$ such that the
Turing program $P_j$ %
computes %
the total function $\omega$ in Definition \ref{domega} if the 
 procedure $Q$, which computes the function $\omega$ according to Theorems \ref{tproc1} and \ref{tproc2},
is applied
to the G\"odel number  $j$
of the Turing program $P_j$ in its second input.
\eth

\bpr
We assume that there is any natural number $j$ such that the 
Turing program $P_j$ computes the total function $\omega$ in order to derive a contradiction.
Our assumption implies that
\beq
\omega(x) = \varphi_j(x) \label{omegax=}
\eeq
for all natural numbers $x$ because
$P_j$ also computes $\varphi_j$ according to Definition \ref{dPi}.
According to the condition in the theorem to be proved
we apply the procedure $Q$ 
to the G\"odel number 
$j$ of the Turing program $P_j$ in its second input.
Because the range of the function $\omega$ %
is a subset of %
$\{ i  |\varphi_i \, {\it total} \}$ %
according to Theorem \ref{trange},
$j$ 
is the G\"odel number
of a
Turing-computable total function whose range is a subset 
of $\{ i  |\varphi_i \, {\it total} \}$, that is,
$j$ is an admissible input
of the procedure $Q$ in Definition \ref{dproc}.
Because the second input variable $j$ of the procedure $Q$ is set,
that is, the condition in the second line of
the procedure $Q$ in Table~\ref{computing} 
is satisfied, 
the third line adds the input-output pair 
$(l,\psi(j))$ %
to the function $\alpha$, that is,
\beq
\alpha(l)=\psi(j), \label{alphapsij}
\eeq
where 
$l = \text{\it least } \{ n \in {\bf N} \, | \ n \notin \text{\it domain } \alpha \}$ %
according to the first line in Table~\ref{computing}.
Because of (\ref{alphapsij}), $l \in \text{\it domain } \alpha$
and
\beq
\omega(l)=\alpha(l)=\psi(j)
\eeq
because of the seventh line in Table~\ref{computing}.
Because of (\ref{omegax=}),
\beq
\omega(l) = \varphi_j(l). \label{omegal=}
\eeq
This implies that $\psi(j)$ is contained in the range of the function $\varphi_j$.
According to Theorem \ref{tprod} the value
 $\psi(j)$ is not contained in the range of the function $\varphi_j$.
Thus, we have derived a contradiction from our original assumption that 
any 
Turing program $P_j$ computes the total function $\omega$.
Therefore, there is no 
Turing program $P_j$,
where $j$ is any natural number,
 that computes the total function $\omega$.
\epr

The condition "if the procedure $Q$ ..." in Theorem  \ref{tomega} %
can be implemented by the following if-then rule:
\br \label{rQ}
If $j$ is the G\"odel number  of a Turing program $P_j$
that computes a total function
whose range is a subset of %
$\{ i  |\varphi_i \, {\it total} \}$, 
then apply the procedure $Q$ in Definition \ref{dproc} %
to $x$ and $j$, where
the value of the input variable $x$ is not set.
\er

If Rule \ref{rQ} is applied, the condition "if the procedure $Q$ ... in its second input" in Theorem  \ref{tomega} 
is satisfied. This implies that there is no
Turing program $P_j$, where $j$ is any natural number, 
computing the total function $\omega$ in Definition \ref{domega}.

The condition in Rule \ref{rQ} may be modified. 
For example, the condition %
may require
a formal proof that %
$P_j$
``computes a total function 
whose range is a subset of %
$\{ i  |\varphi_i \, {\it total} \}$."

\section{Discussion}\label{SDiscussion}
The  proofs of Theorems \ref{tproc1}, \ref{tproc2}, \ref{trange}, and \ref{tomega} in Section \ref{SProc} 
constitute precise evidence
that all elementary operations of the procedure $Q$ in Table \ref{computing} 
are Turing-computable and the total function $\omega$, 
which is computed by the procedure $Q$,
is not Turing-computable.
The theorems are independent of the special productive function $\psi$
of the procedure $Q$ in Table \ref{computing}, that is, 
Theorems \ref{tproc1}, \ref{tproc2}, \ref{trange}, and \ref{tomega} remain valid if the procedure $Q$
in Table \ref{computing} uses any productive function. 
This suggests the following definition:

\bdf \label{dcreative}
A procedure, whose elementary operations are computable by Turing programs,
is called {\it creative} if it computes functions that are not computable by Turing programs.
\edf
According to Theorems \ref{tproc1} and \ref{tproc2} the procedure $Q$ computes the total function $\omega$
in Definition \ref{domega} which is not Turing-computable according to Theorem \ref{tomega}.
This means that the procedure $Q$ is creative in the sense of Definition \ref{dcreative}.

Rule \ref{rQ} in Section \ref{SProc} provides
 G\"odel numbers  $j$ of Turing programs $P_j$
that compute total functions
whose range is a subset of $\{ i  |\varphi_i \, {\it total} \}$, that is, existing information,
in the input of the procedure $Q$
but these   G\"odel numbers  $j$ cannot completely be represented in $Q$
because  the set $\{ i  |\varphi_i \, {\it total} \}$ is productive. 
The procedure $Q\/$ is creative according to Definition \ref{dcreative}
because $Q$ computes the function $\omega$ which is not Turing-computable.
Thus,
Rule \ref{rQ} is an implementation of the following general principle for creative procedures:
\bpri
Existing information, for example, G\"odel numbers  of %
Turing programs,
is provided in the input of procedures
which are creative according to Definition \ref{dcreative}
if the information cannot completely be represented in the procedure.
\epri
The existence principle, in particular, Rule \ref{rQ},
can be implemented physically, that is, in a ``machine". %
Let 
\beq
L(j),    \label{Lj}
\eeq
where $j$ is any natural number, 
be 
formal propositions that stand for the condition
``$j$ is the G\"odel number  of a Turing program $P_j$
that computes a total function
whose range is a subset of %
$\{ i  |\varphi_i \, {\it total} \}$"
in Rule \ref{rQ},
that is,
$L$ in (\ref{Lj}) refers  to a fixed finite string.
The formal proposition $L(j)$ has a physical
representation. Rule \ref{rQ}
is applied if the proposition $L(j)$ in (\ref{Lj}), in particular,
   the G\"odel number  $j$, is generated physically.
The existence principle simply implies that existing information, which is available
physically, is provided in the input of creative procedures such as the procedure $Q$ 
in Table \ref{computing}. %
In view of the results described previously,
this means that the function $\omega$ can neither be modeled by any Turing machine nor
be dealt within a formal system although the computation of $\omega$
can be implemented physically, that is, in a ``machine".

The function $\alpha$ in the procedure $Q$ represents the function $\omega$ in the sense
that the input-output pairs in $\alpha$ are a finite subset of the input-output pairs of $\omega$
 (see  Definition \ref{dproc} %
and Table \ref{computing} in Section \ref{SProc}).
But there is no G\"odel number  for 
the function 
$\omega$, which is computed by %
$Q$,
because input-output pairs $(l, \psi(j))$ and $(x,c)$  may be added to the function $\alpha$
whenever %
$Q$ is executed.

Referring to 
his ``Theorem 2.4, with its corollaries"
Davis \shortcite[pp.\ 121-122]{Davis58} writes:
\bqu
... these results 
really constitute an abstract form of G\"odel's famous incompleteness theorem ... they imply that 
{\em an adequate development of the theory of natural numbers, within a logic L, to the point 
where membership in some given set $Q$ of integers can be adequately dealt with
within the logic ... is possible only if $Q$ happens to be recursively enumerable.}
Hence, non-recursively enumerable sets can, at best, be dealt with in an incomplete manner.
\equ
This implies that the function $\omega$,
which is computed by the procedure $Q$,
cannot be dealt with 
within a logic, that is, a formal system because the range of $\omega$ is not recursively enumerable 
according to Theorem \ref{tomega}.

Let $R$ be any Turing program
that generates any sequence of  G\"odel numbers   
\beq
j_1, j_2, ... \label{j1j2...}
\eeq
of 
Turing programs $P_{j_1}$, $P_{j_2}$, ... computing 
total functions %
whose range is a subset of $\{ i  |\varphi_i \, {\it total} \}$,
that is, $R$ successively generates
the  G\"odel numbers   $j_1$, $j_2$, ... in (\ref{j1j2...}).\footnote{The  sequence of G\"odel numbers $j_1$, $j_2$, ... in Example \ref{Qexseq} in Section \ref{SProc} can be generated by such a Turing program $R$.}
Rule \ref{rQ} applies the procedure $Q$ 
in Table~\ref{computing} 
to any G\"odel number  in (\ref{j1j2...})
as soon as such a G\"odel number  is generated.
Rule \ref{rQ} also processes G\"odel numbers  
that are {\em not} generated by the Turing program $R$,
that is, if the G\"odel number, say $j_k$,
of {\em any} Turing program $P_{j_k}$
computing a total function is generated
whose range is a subset of the set $\{ i  |\varphi_i \, {\it total} \}$,
then Rule \ref{rQ} applies the procedure $Q$ 
in Table~\ref{computing} 
to $j_k$.
Thus, 
according to Theorem \ref{tomega} in Section \ref{SProc}
 there is no  G\"odel number  $j_k$
such that $P_{j_k}$
computes the total function $\omega$ in Definition \ref{domega}.

If the G\"odel numbers  $j$ in the input of Rule \ref{rQ} are restricted to 
the G\"odel numbers   $j_1, j_2, ...$ in (\ref{j1j2...}),
 which are generated by the Turing program $R$,
the range of the total function $\omega$ in Theorem \ref{tomega}
is recursively enumerable.
But such a restriction restricts 
the input of Rule \ref{rQ}
to a recursively enumerable subset of
the set $\{ i  |\varphi_i \, {\it total} \}$ 
 and thus restricts
the generality 
of Rule \ref{rQ} because the set
 $\{ i  |\varphi_i \, {\it total} \}$ is productive. 
This means that
 the inputs of 
Rule \ref{rQ} cannot be restricted to a recursively enumerable subset 
of $\{ i  |\varphi_i \, {\it total} \}$
because a productive function $\psi$ for  $\{ i  |\varphi_i \, {\it total} \}$
could be applied to a  G\"odel number  $j$ 
of a Turing program generating this subset and thus produce the 
 G\"odel number  $\psi(j)$ of a
Turing-computable function that is  contained in $\{ i  |\varphi_i \, {\it total} \}$ but not in this subset.
Thus, the range of the total function $\omega$ in Theorem \ref{tomega}
is not recursively enumerable because the inputs of 
Rule \ref{rQ} cannot be restricted to a recursively enumerable subset 
of the productive set $\{ i  |\varphi_i \, {\it total} \}$.

Rogers 
\shortcite[pp.\ 10--11]{Rogers87} discusses
``the problem of getting a satisfactory
 [formal] %
characterization of algorithm and algorithmic function"
because the application of diagonalization
to a list of total algorithmic functions yields
a  total algorithmic function that is not contained in the list
(see the proof of Theorem \ref{tprod} in Section \ref{SProc}).
Rogers 
\shortcite[pp.\ 11--12]{Rogers87} writes:
\bqu
{\it We can avoid the diagonalization difficulty 
by allowing sets
of instructions for nontotal partial functions as  well as 
for total functions.}
...
The approach taken by way of partial functions is, in essence,
the approach taken by Kleene ..., Church ..., Turing [1936]
and others in the 1930's. 
\equ
Referring to ``the concept of general recursiveness
(or Turing's computability)"
G\"odel \shortcite[p.\ 84]{Goedel46}
writes:
\bqu
... with this concept one has for the first time succeeded
in giving an absolute definition of an interesting
epistemological notion, i.e., one not depending
on the formalism chosen.
In all other cases treated previously, such as demonstrability %
... it is clear that the one %
 obtained
is not the one looked for.
...
By a kind of miracle ... the diagonal procedure does not lead outside
the defined notion.
\equ
Thus, 
there is only a formal characterization of 
a single ``interesting
epistemological notion", that is,
the Turing-computable  partial functions.
In contrast, creative procedures 
provide a framework
for an investigation of other ``notions"
such as 
 Turing-computable total functions.
This framework is not subject to a 
``diagonalization difficulty"
but uses diagonalization to
investigate such ``notions" which
cannot be captured by formal systems.

Discussing the formalization of a theory which results in a formal system, Kleene \shortcite[p.\ 64]{Kleene52} 
writes: 
\bqu
Metamathematics must study the
formal system as a system of symbols, etc.\ which are considered wholly objectively. This means 
simply that those
symbols, etc.\ are themselves the ultimate objects, and are not being used to refer to something other than
themselves. The metamathematician looks at them, not through and beyond them; thus they are objects without interpretation or meaning.
\equ
Referring to his undecidable formula ${\rm A}_p(\textbf{\textit{p}})$ in G\"odel's 
incompleteness theorem 
Kleene \shortcite[p.\ 426]{Kleene52} writes:
\bqu
... if we suppose the number-theoretic formal system
to be consistent, we can recognize that ${\rm A}_p(\textbf{\textit{p}})$ 
is true by taking into view the structure of that system as a whole,
though
we cannot recognize the truth of  ${\rm A}_p(\textbf{\textit{p}})$ 
by use only of the principles of inference formalized in that system,
i.e. not $ \vdash {\rm A}_p(\textbf{\textit{p}})$.\footnote{The expression
``not $ \vdash  {\rm A}_p(\textbf{\textit{p}})$" %
in  Kleene \shortcite{Kleene52} means
that the undecidable formula ${\rm A}_p(\textbf{\textit{p}})$ 
in G\"odel's theorem is not provable in the formal
system.} 
\equ
Thus, G\"odel's theorem requires a reference to the (incomplete) formal ``system
as a whole"
which cannot be achieved within the formal system itself
because, in our interpretation, the formal system  cannot take ``into view the structure of that system as a whole",
that is, ``wholly objectively".
In particular, a formal system represented by a Turing program 
cannot take ``into view the structure of that"  Turing program ``as a whole",
that is, ``wholly objectively",
in the sense that the Turing program, which produces
a (recursively enumerable)
subset of a productive set,
 cannot refer to itself and thus capture 
the result of applying a productive partial function
for the productive set to
its own G\"odel number. %

The assumption
that all reasoning processes can 
be modeled by a Turing program (see Section \ref{SIntr}) immediately yields a contradiction
if the following if-then rule, which is also an implementation of the existence principle, is used:
\br \label{rpsi}
If $j$ is the G\"odel number  of a Turing program $P_j$
that computes a total function
whose range is a subset of %
$\{ i  |\varphi_i \, {\it total} \}$, 
then $\psi(j)$,
where $\psi$ is a productive partial function for $\{ i  |\varphi_i \, {\it total} \}$, 
is the G\"odel number of a Turing-computable total function,
that is,  $\varphi_{\psi(j)}$ is a total function.
\er
\noindent
Rule \ref{rpsi} is an immediate implication 
of Theorem \ref{tprod} in Section \ref{SProc}.
A pseudo-code for  Rule \ref{rpsi} 
is
\beq
\textit{if } \; L(j) \textit{ then } \; T(\psi(j)),  \label{LT}
\eeq
where $L(j)$ is a
formal proposition which stands for the condition
``$j$ is the G\"odel number  of a Turing program $P_j$ ... a subset of $\{ i  |\varphi_i \, {\it total} \}$"
in Rule \ref{rpsi}
and
$T(j)$ is a
formal proposition which stands for the consequent
``$\psi(j)$ ...
is the G\"odel number of a Turing-computable total function"
in Rule \ref{rpsi},
that is,
$L$ and $T$ in (\ref{LT}) refer  to  fixed finite strings.
If we assume 
that all reasoning processes can 
be modeled by a Turing program, say $P_k$,
then $P_k$ computes
a total function
whose range is 
the subset of %
$\{ i  |\varphi_i \, {\it total} \}$
containing all members of $\{ i  |\varphi_i \, {\it total} \}$
that are generated  by $P_k$.
Let $j$ be the G\"odel number $j$  of a Turing program $P_j$
that computes this total function.
The use of Rule \ref{rpsi}, which is an implementation
of the existence principle,
implies that a formal representation (\ref{LT})
of Rule \ref{rpsi}
and a formal representation $L(t)$ of its 
condition are contained in the Turing program $P_k$
which is assumed to model all reasoning processes.
The application of (\ref{LT}) to $L(t)$ by 
the Turing program $P_k$ yields the consequent
 $T(\psi(j))$ in (\ref{LT}).
 Thus,  we have a contradiction
 because, according to Theorem \ref{tprod}   
in Section \ref{SProc}, 
 $\psi(j)$ 
 is not contained in the range of $\varphi_j$
 which contains all members of $\{ i  |\varphi_i \, {\it total} \}$
that are generated  by $P_k$. %
These considerations also apply to the ``reasoning processes" 
of a ``machine" or ``robot"
because, as described above,
the existence principle, in particular, Rules  \ref{rQ} and \ref{rpsi}, can be implemented physically.

As discussed above, 
a restriction of the input of Rule \ref{rQ} 
to a recursively enumerable subset of $\{ i  |\varphi_i \, {\it total} \}$
restricts
the generality 
of Rule \ref{rQ}.
The use of Rule \ref{rpsi},
 which is an implementation of the existence principle,  also implies that 
 such a restriction 
restricts the generality of Rule \ref{rQ}
because the application of 
Rule \ref{rpsi} to the G\"odel number $j$ of 
a Turing-computable total function
whose range is 
the subset of %
$\{ i  |\varphi_i \, {\it total} \}$
yields the 
G\"odel number $\psi(j)$ of 
a Turing-computable total function
that is not contained in 
this subset but satisfies the condition in Rule \ref{rQ}.
A proof that $j$ is
the G\"odel number of 
a Turing-computable total function
whose range is 
the subset of %
$\{ i  |\varphi_i \, {\it total} \}$
can be extended to a proof that 
$\psi(j)$ is the G\"odel number of 
a Turing-computable total function
and not contained in 
this subset because $\psi$ is a productive partial function. 

No %
formal system can capture its own existence in the sense that
 every Turing program representing a formal system and generating (recursively enumerable)
subsets of productive sets cannot generate the result of applying productive partial functions
for the productive sets to
the G\"odel numbers  of Turing programs
producing these %
subsets
(see Theorem \ref{tprod} in Section \ref{SProc} and the quotation
from Davis \shortcite[pp.\ 121-122]{Davis58} above). 
This implies that a Turing program generating a (recursively enumerable)
subset of a productive set
cannot contain
a reference to its own G\"odel number 
because a productive function for the productive set
could be applied to such a reference
such that the original Turing  program
could contain the output of the productive function.
The incompleteness of formal systems is %
 ``overcome"
by the existence principle which can be implemented in rules such as Rule \ref{rQ}. %
Roughly speaking, this principle 
implies that existing information %
is provided in
 the input of creative procedures %
and thus ``overcomes" the incompleteness of formal systems
and the limits of Turing's computability.\footnote{Post \shortcite[p.\ 423]{Post65} writes: 
``What we must now do is to isolate the creative germ in the thinking process."}
Thus, the second input variable $j$ in the procedure $Q$ in Table \ref{computing}
is required because formal systems including Turing programs cannot
refer to existing information such as their own G\"odel numbers.
\vspace*{-.09cm}

\section{Related Work}  \label{SRelated}

Church's \shortcite[pp.\ 90, 100-102]{Church35} thesis\footnote{The term {\em Church's thesis} is due to
Kleene \shortcite[p.\ 274]{Kleene43} (see Kleene \shortcite[pp.\ 300, 317]{Kleene52}).}
 states that every effectively calculable function
is general recursive, that is, computable by a Turing machine (see 
Kleene \shortcite[pp.\ 300--301, 317--323]{Kleene52}).
Since ``effective calculability" is an intuitive concept,
the thesis cannot be proved (see Kleene 
\shortcite[p.\ 317]{Kleene52}).\footnote{In his article 
``Why G\"odel Didn't Have Church's Thesis"
Davis \shortcite[p.\ 22, footnote 26]{Davis82}
writes: ``We are not concerned here with attempts to distinguish
'mechanical procedures' (to which Church's thesis is held to apply)
from a possible broader class of 'effective procedures' ..."}

Referring to G\"odel's \shortcite{Goedel31} incompleteness theorem and
Church's \shortcite[pp.\ 90, 100-102]{Church35}  identification of effective
calculability with recursiveness,
Post \shortcite[p.\ 291, footnote 8]{Post36}
writes:
\bqu
``Actually the work already done by Church and others carries this identification 
considerably beyond the working hypothesis stage. But to mask this identification
under a definition hides the fact that 
a fundamental discovery in the limitations of the 
mathematicizing power of Homo Sapiens has been made and blinds us to 
the need of continual verification." 
\equ
Thus, Post calls for a ``continual verification" of Church's thesis
because of the incompleteness of formal systems (see Section \ref{SIntr}).
Referring to Principia Mathematica (see G\"odel
\shortcite{Goedel31}) and his normal systems,
 Post \shortcite[p.\ 408]{Post65}
writes (see Section \ref{SIntr}):
\bqu
... for full generality a complete analysis would have to be made
of all possible ways in which the human mind could set up
finite processes for generating sequences.
\equ
In our view, the existence principle 
can be used by the ``human mind"
because existing information, for example,
a Turing machine representing a formal system, must be
a constituent of the ``human mind", that is, 
a constituent of reasoning.
Thus, the
``human mind" can apply 
an implementation of the existence principle, for example, Rule \ref{rQ},
and the procedure 
$Q$ in Table \ref{computing}
in Section \ref{SProc} 
to ``set up
finite processes for generating sequences"
which cannot be computed by any Turing program 
according to Theorem \ref{tomega}.
This means that a %
formal system
cannot deal with a fundamental aspect of reasoning 
because it cannot refer to its own existence.

The function $\omega$, which is computed by the 
procedure $Q$ in Table~\ref{computing} in Section \ref{SProc}, can 
be regarded as effectively calculable
because the elementary operations of $Q$ 
are Turing-computable according to Theorem~\ref{tproc1}
and Rule \ref{rQ}, which uses the procedure $Q$, can be implemented physically.

The condition "if the procedure $Q$ ..." in Theorem  \ref{tomega} 
is satisfied if Rule \ref{rQ} in Section \ref{SDiscussion}
is applied. Rule \ref{rQ} 
is an implementation of the existence
principle, which states that
existing information, for example,
 G\"odel numbers 
of Turing-computable total functions
whose range is a subset of all Turing-computable total functions,
that is, recursively enumerable subsets of a productive set,
is provided in the input of creative procedures
which contain a productive function for the productive set.
Thus, Church's thesis is not valid if the existence principle is 
applied to recursively enumerable subsets of productive sets
and suitable procedures.

Church \shortcite[pp.\ 90, 102]{Church35} presents his thesis as a ``definition
of effective calculability" and proposes a second definition
of effective calculability:
\bqu
... (2) by defining a function $F$ (of positive integers) to be effectively calculable
if, for every positive integer $m$,
there exists a positive integer $n$
such that $F(m)=n$ is a provable theorem.
\equ
If we require for every input $j$ of the procedure $Q$ in Table~\ref{computing}
a formal proof that the Turing program $P_j$ computes a total
function (see the modification of Rule \ref{rQ} in Section \ref{SProc}),
then, for every natural number (positive integer) $x$
in the input of $Q$, which computes the function $\omega(x)$,
there exists a natural number $y$
such that $\omega(x)=y$ is a provable theorem in some 
formal system, say $S_x$.
Such a formal system also exists 
for any finite set of natural numbers $x$
in the input of $Q$.
But, because of Theorem \ref{tomega} in Section \ref{SProc}, there exists no formal system $S$
such that $\omega(x)=y$, where $y$ is a natural number, is a provable theorem in $S$
for all natural numbers $x$.\footnote{Here, we implicitly assume that
the formal system $S$
is represented by a Turing machine and a G\"odel number  for
the recursively enumerable subset of $\{ i  |\varphi_i \, {\it total} \}$
produced by $S$ is generated
such that Theorem \ref{tomega} and Rule \ref{rQ} are applicable.} 
Roughly speaking,  Theorem \ref{tomega} implies that
the formal systems $S_x$ cannot be unified into a single formal system $S$.

In a
letter of June 8, 1937, to Pepis
Church wrote (see
Sieg \shortcite[pp.\ 175--176]{Sieg97}): 
\bqu
... if a
numerical function $f$ is effectively calculable then for every positive
integer $a$ there must exist a positive integer $b$ such that a valid proof
can be given of the proposition $f(a) = b$ ...

Therefore to discover a function which was effectively calculable
but not general recursive would imply discovery of an utterly new
principle of logic, not only never before formulated, but never before
actually used in a mathematical proof - since all extant mathematics is
formalizable within the system of Principia [Mathematica], or at least within one of
its known extensions. %
 \equ
As far as we know
the existence principle was 
``never before
actually used in a mathematical proof".

The function $\alpha$ in the procedure $Q$ in Definition \ref{dproc} and Table \ref{computing},
which is represented as a set of input-output pairs,
is a subset of the input-output pairs of the function $\omega$. %
$Q$ %
returns $\alpha(x)$ as the output of $\omega(x)$ for any natural number $x$ in the first input of $Q$,
that is, the input of $\omega$. 
Nevertheless, the set $\alpha$ of input-output pairs, which is the empty set $\emptyset$
before the first execution of %
$Q$,
is finite at every point in time.
Obviously, each set $\alpha$ of input-output pairs 
can be generated by a Turing machine. %
Thus,
 the number of states of Turing programs
generating %
$\alpha$
is finite ``at each stage" 
of its development %
but according to Theorem \ref{tomega} 
there is 
no G\"odel number 
of a Turing program computing $\omega$, %
that is, %
$\alpha$
 ``at each stage" of its development %
(see G\"odel \shortcite[p.\ 306]{Goedel72a} and
Section \ref{SIntr}).
This means that the number of states of the Turing machines
generating $\alpha$ %
will not be ``confused" by an ``infinity of states of mind" because this number of states is finite ``at each stage"
of the development of $\alpha$, that is, the development of $\omega$ %
 (see Turing \shortcite[pp.\ 249--250]{Turing36} and
Section \ref{SIntr}).\footnote{We suppose Turing's \shortcite[pp.\ 249--250]{Turing36} infinity is an infinity according to a mathematical definition. For example, a definition states that a set is {\em infinite} if it is not finite.
Another definition states that a set $A$ is {\em Dedekind-infinite} if 
some proper subset $B$ of $A$ is equinumerous to $A$, that is,
there is a bijection (one-to-correspondence) between $A$ and $B$.
In Zermelo-Fraenkel set theory this definition is equivalent to
the condition that a set $A$ is infinite if there is a one-to-one correspondence between 
all natural numbers and a subset of $A$. 
The number of states of the Turing machines
generating $\alpha$ 
is not finite %
and
seems to be ``unbounded" %
because there is no Turing machine generating $\alpha$ %
 ``at each stage" of its development 
if Rule \ref{rQ} is applied. 
Hilbert \shortcite[pp.\ 183-186]{Hilbert26} writes: "... the {\it infinite}, as that concept is used in mathematics, 
has never been completely clarified ... the infinity in the sense of an infinite totality, where we still find it used in deductive methods, is an illusion. ... no other {\it concept} needs {\it clarification} more it does."}

Referring to Church's thesis and G\"odel's ``mental procedures" Kleene \shortcite[pp.\ 493, 494]{Kleene87} writes:
\bqu
For, in the idea
of ``effective calculability" or of an ``algorithm" as I understand it, 
it is essential that all of the infinitely many 
calculations
... %
are performable ... %
by following a set of instructions fixed in
advance of all the calculations. If the Turing machine representation is used,
this includes there being only a finite number of ``internal machine configurations",
corresponding to a finite number of a human computer's mental states.
...
 As Turing ... says ..., ``If we admitted an infinity of states of mind, some of them will be
`arbitrarily close' and will be confused." Hardly appropriate for keeping things
straight digitally!

... an effective (finitely describable) procedure from the beginning,
coming under
the Church-Turing thesis. 
 \equ
In our view, the descriptions of the procedure $Q$ in Section \ref{SProc}
and Rule \ref{rQ} in Section \ref{SDiscussion}
are ``fixed"  
at the beginning and ``finite"  at every point in time but the function $\alpha$ in the procedure $Q$
develops in the course of time
because it depends on
the input variables $x$ and $j$ of $Q$, that is,
the functions $\alpha$ and $\omega$ cannot be described ''in advance"
because the set of the G\"odel numbers of the Turing-computable total
functions is productive.
As described above,
the number of states of Turing machines
generating the function $\alpha$ 
will not be ``confused" by an ``infinity of states of mind" because this number of states is finite ``at each stage"
of the development of $\alpha$, that is, the development of $\omega$ 
 (see Turing \shortcite[pp.\ 249--250]{Turing36} and
Section \ref{SIntr}).

The proof of Theorem \ref{tprod} in Section \ref{SProc}
states that the G\"odel number 
$\psi(j)$ of 
the total function $\varphi_{\psi(j)}$ 
is not contained in the range of 
$\varphi_j$, where
$j$ is any natural number such that $\varphi_j$ is a total function with 
$range \, \varphi_j \subseteq \{ i  |\varphi_i \, {\it total} \}$.
In particular, the proof
states that the total function $\varphi_{\psi(j)}$ 
 is different from 
$\varphi_{\varphi_j(n)}$ for all natural numbers $n$.
This means that the productive function $\psi$
produces not only a new G\"odel number $\psi(j)$,
which is different from 
$\varphi_j(n)$
for all natural numbers $n$,
but also a new total function $\varphi_{\psi(j)}$,
which
 is different from all total functions
$\varphi_{\varphi_j(n)}$ for all natural numbers $n$.

G\"odel \shortcite[p.\ 306]{Goedel72a} writes (see Section \ref{SIntr}):
\bqu
... {\em mind ... %
is not static, but constantly developing}, ... 
\equ
If creative procedures and the existence principle are used,
{\it ''mind"\/} is
{\it ''constantly developing"} in the sense that it
produces new structures %
from existing structures.
For example, the use of Rule \ref{rQ},
which is an implementation of the existence principle,
produces a 
new G\"odel number $\psi(j)$ in the creative procedure $Q$ whenever 
any natural number $j$ such that $\varphi_j$ is a total function with 
$range \, \varphi_j \subseteq \{ i  |\varphi_i \, {\it total} \}$
is generated.
As described above, 
 $\varphi_{\psi(j)}$ is a new total function
which
 is different from all total functions
$\varphi_{\varphi_j(n)}$ for all natural numbers $n$.\footnote{All existing structures
in a creative procedure are called ``reflection base".
The repeated application of the existence principle,
that is, the repeated application of a creative procedure
to information in its reflection base,
 can be regarded as a feedback
 process which produces more and more powerful structures.
 A first step towards the implementation of a  creative procedure
 is described in Ammon \shortcite{Ammon88},
 Ammon \shortcite{Ammon92a}, and Ammon \shortcite{Ammon93}.
 These experiments suggest 
 that creative procedures are 
 a self-developing process which can start from any universal
 programming language.
 Their structure can be regarded as a web 
of %
concepts and methods which %
are called ``analytical spaces" and 
cannot be characterized formally. 
This is plausible because formal systems are restricted
to Turing-computable partial functions, that is, recursively enumerable sets
(see the quotations from Davis \shortcite[pp.\ 121-122]{Davis58},
Rogers \shortcite[pp.\ 11--12]{Rogers87}, and G\"odel \shortcite[p.\ 84]{Goedel46}
 in Section~\ref{SDiscussion}).
 }

Lucas \shortcite{Lucas61} argues %
that mind cannot be modeled by %
 a Turing
machine 
because he {\em knows} that the %
undecidable
proposition in G\"odel's theorem 
is true (see Shapiro \shortcite[pp.\ 273-274]{Shapiro98}).
Putnam points out that Lucas cannot prove the prerequisite of consistency
in G\"odel's 
theorem (see Shapiro \shortcite[pp.\  282-284]{Shapiro98}).
The procedure $Q$ %
and Rule \ref{rQ} in Section \ref{SProc} %
 can be executed
by a human and by %
 a ``machine" to compute the function $\omega$. 
According to %
Theorem %
\ref{tomega}
there is no Turing machine computing %
$\omega$.
 Thus, an implementation of the existence principle, which merely uses the existence
of a formal system, that is, a Turing machine itself, 
``overcomes" its incompleteness, that is, the limits of the Turing 
machine.\footnote{G\"odel \shortcite[pp.\ 71--72]{Goedel65} writes: ``... due
to A.M.\ Turing's work, a precise and unquestionably adequate definition
of the general concept of formal system
can now be given, ... Turing's work gives an analysis of the concept
of 'mechanical procedure' (alias 'algorithm' ...). This concept is shown to be 
equivalent with that of a 'Turing machine'. A formal system can simply be defined
to be any mechanical procedure for producing formulas, called provable formulas.}
Referring to Penrose \shortcite{Penrose90},
Davis \shortcite[p.\ 611]{Davis93} writes:
\bqu
However, it [G\"odel's theorem] is a quite ordinary sentence of elementary number theory and can be proved with no
difficulty whatever in any formal system adequate for elementary number theory, such as
for example Peano arithmetic. Note that this powerful form of G\"odel's theorem applies
uniformly to any formalism whatever. 
\equ
Although %
G\"odel's theorem applies to any (given) ``formalism"\footnote{Originally,
G\"odel
\shortcite{Goedel31} proved his theorem for the ``formalism" of
``Principia Mathematica and related systems".}
and can be proved in a ``formal system",
no proof of G\"odel's theorem in any 
``formal system" can apply to this ``formal system" itself
because, as described above, no ``formal system" can refer to itself, that is,
to its own existence.
Rather, a proof of G\"odel's theorem for a ``formalism", that is, ``formal system", say $S_1$,
requires another extended ``formal system", say $S_2$,
which refers to $S_1$,
a proof of G\"odel's theorem for $S_2$,
requires another extended ``formal system", say $S_3$, 
which refers to $S_2$,
and so on.\footnote{In particular, a reference to ``any formalism whatever"
cannot be formalized.
For example, a Turing program generating a 
(recursively enumerable) subset of a productive set,
is a ``formalism", that is, a formal system.
A formal reference to all these Turing programs does not exist
because productive sets are not recursively enumerable.
Mendelson \shortcite[p. 253]{Mendelson64} writes that
there are $2^{\aleph_0}$ productive sets, that is, the cardinality of the 
productive sets corresponds to the cardinality of the continuum. 
This also implies that 
 a reference to ``any formalism whatever"
cannot be formalized.
}
Thus, there is no proof in any formal system
showing that G\"odel's theorem applies
``to any formalism whatever". 
Therefore, the ``insight" that  
 G\"odel's theorem applies
``to any formalism whatever"
requires 
a new principle such as the existence principle which
 cannot be formalized although it 
 can be implemented physically.

\section{Conclusion}  

We described a procedure that computes a total function
whose range can contain members of a productive set.
The elementary operations of the procedure are Turing-computable.
We proved that there is no Turing program
computing this total function if the existence principle is
used which implies that existing recursively enumerable subsets of the productive set
are provided in the second input of the procedure. 
The existence principle can be implemented in rules
which can be executed by humans and ``machines".
Roughly speaking, a formal system cannot contain a reference to
itself and an extension of the recursively enumerable sets that it deals with by productive
functions.
In view of the existence principle, this means that a formal system 
cannot deal with a fundamental aspect of reasoning, that is, it cannot capture its own existence.
Church's thesis is not valid if the existence principle is
applied to recursively enumerable subsets of productive sets
and suitable procedures.
We defined creative procedures which 
compute functions that are not computable by Turing machines
and argued that creative procedures model an aspect of reasoning that
cannot be modeled by Turing machines.

\bigskip \medskip %
\noindent
{\bf Acknowledgments}. The author wishes to thank colleagues, in  particular,
Sebastian Stier and Andreas Keller, for helpful comments on earlier versions of 
this paper.

\bibliography{Effective}

\begin{thebibliography}{}

\bibitem[\protect\citeauthoryear{{Ammon}}{1988}]{Ammon88}
K.~{Ammon}.
\newblock The automatic acquisition of proof methods.
\newblock In {\em National Conference on Artificial Intelligence, St. Paul},
  San Mateo, Calif., 1988. Morgan Kaufmann.

\bibitem[\protect\citeauthoryear{{Ammon}}{1992}]{Ammon92a}
K.~{Ammon}.
\newblock Automatic proofs in mathemetical logic and analyis.
\newblock In {\em 11th International Conference on Automated Deduction,
  Saratoga Springs}, pages 4--19, Berlin, 1992. Springer.

\bibitem[\protect\citeauthoryear{{Ammon}}{1993}]{Ammon93}
K.~{Ammon}.
\newblock An automatic proof of {G}{\"o}del's incompleteness theorem.
\newblock {\em Artificial Intelligence}, 61(2):291--306, 1993.

\bibitem[\protect\citeauthoryear{{Church}}{1965}]{Church35}
A.~{Church}.
\newblock An unsolvable problem of elementary number theory.
\newblock In M.~{Davis}, editor, {\em The Undecidable}, pages 89--107. Raven
  Press, New York, 1965.
\newblock Reprinted from {\it The American Journal of Mathematics}, vol. 58,
  pp. 345-363 (1936).

\bibitem[\protect\citeauthoryear{{Davis}}{1982a}]{Davis58}
M.~{Davis}.
\newblock {\em Computability and Unsolvability}.
\newblock Dover, New York, 1982.

\bibitem[\protect\citeauthoryear{{Davis}}{1982b}]{Davis82}
M.~{Davis}.
\newblock Why {G\"o}del didn't have {C}hurch's thesis.
\newblock {\em Information and Control}, 54:3--24, 1982.

\bibitem[\protect\citeauthoryear{{Davis}}{1993}]{Davis93}
M.~{Davis}.
\newblock How subtle is {G\"o}del's theorem? {M}ore on {R}oger {P}enrose.
\newblock {\em Behavioral and Brain Sciences}, 16:611--612, 1993.

\bibitem[\protect\citeauthoryear{{Dekker}}{1955}]{Dekker55}
J.~{Dekker}.
\newblock Productive sets.
\newblock {\em Transactions of the American Mathematical Society},
  78(1):129--149, 1955.

\bibitem[\protect\citeauthoryear{{G\"odel}}{1965a}]{Goedel31}
K.~{G\"odel}.
\newblock On formally undecidable propositions of the {P}rincipia {M}athematica
  and related systems {I}.
\newblock In M.\ Davis, editor, {\em The Undecidable}, pages 4--38. Raven
  Press, New York, 1965.
\newblock The original German title is: \"Uber formal unentscheidbare S\"atze
  der Principia Mathematica und verwandter Systeme I. {\it Monatshefte f\"ur
  Mathematik und Physik}, vol. 38 (1931), pp. 173-198.

\bibitem[\protect\citeauthoryear{{G\"odel}}{1965b}]{Goedel65}
K.~{G\"odel}.
\newblock On undecidable propositions of formal mathematical systems -
  {POSTSCRIPTUM}.
\newblock In M.\ Davis, editor, {\em The Undecidable}, pages 39--74. Raven
  Press, New York, 1965.

\bibitem[\protect\citeauthoryear{{G\"odel}}{1965c}]{Goedel46}
K.~{G\"odel}.
\newblock Remarks before the princeton bicentennial conference on problems in
  mathematics.
\newblock In M.\ Davis, editor, {\em The Undecidable}, pages 84--88. Raven
  Press, New York, 1965.

\bibitem[\protect\citeauthoryear{{G\"odel}}{1990}]{Goedel72a}
K.~{G\"odel}.
\newblock G{\"o}del 1972a: Some remarks on the undecidability results.
\newblock In S.~Feferman et~al., editors, {\em Collected Works: Publications
  1938-1974}, volume~2. Oxford University Press, New York, 1990.

\bibitem[\protect\citeauthoryear{{Hilbert}}{1983}]{Hilbert26}
D.~{Hilbert}.
\newblock On the infinite.
\newblock In P.~{Benacerraf} and H.~{Putnam}, editors, {\em Philosophy of
  Mathematics: Selected Readings}, pages 183--201. Cambridge University Press,
  Cambridge, 1983.
\newblock Translated by Erna Putnam and Gerald J.\ Massey from {\it
  Mathematische Annalen} (Berlin) vol. 95 (1926), pp. 161-190.

\bibitem[\protect\citeauthoryear{{Kleene}}{1952}]{Kleene52}
S.~C. {Kleene}.
\newblock {\em Introduction to Metamathematics}.
\newblock North-Holland, Amsterdam, 1952.

\bibitem[\protect\citeauthoryear{{Kleene}}{1965}]{Kleene43}
S.~C. {Kleene}.
\newblock Recursive predicates and quantifiers.
\newblock In M.\ Davis, editor, {\em The Undecidable}, pages 255--287. Raven
  Press, New York, 1965.
\newblock Reprinted from {\it Transactions of the American Mathematical
  Society}, Volume 53 (1943), No. 1, pages 41--73.

\bibitem[\protect\citeauthoryear{{Kleene}}{1987}]{Kleene87}
S.~C. {Kleene}.
\newblock Reflections on {C}hurch's thesis.
\newblock {\em Notre Dame Journal of Formal Logic}, 28(4):490--498, 1987.
\newblock Available at \url{http://projecteuclid.org/euclid.ndjfl/1093637645}
  (viewed Jan. 7, 2011).

\bibitem[\protect\citeauthoryear{{Lucas}}{1961}]{Lucas61}
J.~R. {Lucas}.
\newblock Minds, machines, and {G\"o}del.
\newblock {\em Philosophy}, 36:112--137, 1961.

\bibitem[\protect\citeauthoryear{{Mendelson}}{1964}]{Mendelson64}
E.~{Mendelson}.
\newblock {\em Introduction to Mathematical Logic}.
\newblock Van Nostrand Reinhold Company, New York, 1964.

\bibitem[\protect\citeauthoryear{{Penrose}}{1990}]{Penrose90}
R.~{Penrose}.
\newblock Author's response: the nonalgorithmic mind.
\newblock {\em Behavioral and Brain Science}, 13:692--705, 1990.

\bibitem[\protect\citeauthoryear{{Post}}{1944}]{Post44}
E.~{Post}.
\newblock Recursively enumerable sets of positive integers and their decision
  problems.
\newblock {\em Bulletin of the American Mathematical Society}, 50(5):284--316,
  1944.
\newblock Available at
  \url{http://www.ams.org/journals/bull/1944-50-05/S0002-9904-1944-08111-1}
  (viewed Dec. 30, 2011).

\bibitem[\protect\citeauthoryear{{Post}}{1965a}]{Post65}
E.~{Post}.
\newblock Absolutely unsolvable problems and relatively undecidable
  propositions - account of an anticipation.
\newblock In M.~{Davis}, editor, {\em The Undecidable}, pages 338--433. Raven
  Press, New York, 1965.

\bibitem[\protect\citeauthoryear{{Post}}{1965b}]{Post36}
E.~{Post}.
\newblock Finite combinatory processes. {F}ormulation {I}.
\newblock In M.~{Davis}, editor, {\em The Undecidable}, pages 289--291. Raven
  Press, New York, 1965.
\newblock Reprinted from {\em The Journal of Symbolic Logic}, vol. 1 (1936),
  pp.\ 103-105.

\bibitem[\protect\citeauthoryear{{Rogers}}{1987}]{Rogers87}
H.~{Rogers}.
\newblock {\em Theory of Recursive Functions and Effective Computability}.
\newblock The MIT Press, Cambridge, 1987.

\bibitem[\protect\citeauthoryear{{Shapiro}}{1998}]{Shapiro98}
S.~{Shapiro}.
\newblock Incompleteness, mechanism, and optimism.
\newblock {\em The Bulletin of Symbolic Logic}, 4(3):273--302, September 1998.
\newblock Available at \url{http://www.math.ucla.edu/~asl/bsl/0403-toc.htm}
  (viewed Jan. 1, 2012).

\bibitem[\protect\citeauthoryear{{Sieg}}{1997}]{Sieg97}
W.~{Sieg}.
\newblock Step by recursive step: Church's analysis of effective calculability.
\newblock {\em The Bulletin of Symbolic Logic}, 3(2):154--180, 1997.
\newblock Available at \url{http://www.math.ucla.edu/~asl/bsl/0302-toc.htm}
  (viewed Jan. 1, 2011).

\bibitem[\protect\citeauthoryear{{Soare}}{1978}]{Soare78}
R.~I. {Soare}.
\newblock Recursively enumerable sets and degrees.
\newblock {\em The Bulletin of American Mathematical Society},
  84(6):1149--1181, 1978.
\newblock Available at
  \url{http://www.ams.org/bull/1978-84-06/S0002-9904-1978-14552-2} (viewed Dec.
  31, 2011).

\bibitem[\protect\citeauthoryear{{Turing}}{1936}]{Turing36}
A.~M. {Turing}.
\newblock On computable numbers, with an application to the
  {E}ntscheidungsproblem.
\newblock In {\em Proceedings of the London Mathematical Society}, volume~42 of
  {\em series 2}, pages 230--265, 1936.

\bibitem[\protect\citeauthoryear{{Turing}}{1939}]{Turing39}
A.~M. {Turing}.
\newblock Systems of logics based on ordinals.
\newblock In {\em Proceedings of the London Mathematical Society}, volume~45 of
  {\em series 2}, pages 161--228, 1939.
\newblock Available at \url{http://www.turingarchive.org/browse.php/B/15}
  (viewed Apr. 13, 2011).

\end{thebibliography}
\bibliographystyle{named} %

\end{document}